\documentclass[preprint,12pt]{elsarticle}

\usepackage{amssymb}
\usepackage{amsmath}

\usepackage{lineno}

\usepackage[super]{nth}

\usepackage{booktabs}

\journal{Computer Vision and Image Understanding}

\begin{document}

\begin{frontmatter}



\title{Generative Fields: Uncovering Hierarchical Feature Control for StyleGAN via Inverted Receptive Fields} 


\author[label1]{Zhuo He} 
\author[label1]{Paul Henderson} 
\author[label1]{Nicolas Pugeault} 

\affiliation[label1]{organization={University of Glasgow},
            addressline={18 Lilybank Gardens}, 
            city={Glasgow},
            postcode={G12 8RZ}, 
            country={United Kingdom}}

\begin{abstract}
StyleGAN has demonstrated the ability of GANs to synthesize highly-realistic faces of imaginary people from random noise. One limitation of GAN-based image generation is the difficulty of controlling the features of the generated image, due to the strong entanglement of the low-dimensional latent space. Previous work that aimed to control StyleGAN with image or text prompts modulated sampling in $\mathcal{W}$ latent space, which is more expressive than $\mathcal{Z}$ latent space. However, $\mathcal{W}$ space still has restricted expressivity since it does not control the feature synthesis directly; also the feature embedding in $\mathcal{W}$ space requires a pre-training process to reconstruct the style signal, limiting its application.
This paper introduces the concept of ``generative fields” to explain the hierarchical feature synthesis in StyleGAN, inspired by the receptive fields of convolution neural networks (CNNs). Additionally, we propose a new image editing pipeline for StyleGAN using generative field theory and the channel-wise style latent space $\mathcal{S}$, utilizing the intrinsic structural feature of CNNs to achieve disentangled control of feature synthesis at synthesis time. 
\end{abstract}

\begin{graphicalabstract}

\centering
\includegraphics[width=1.0\textwidth]{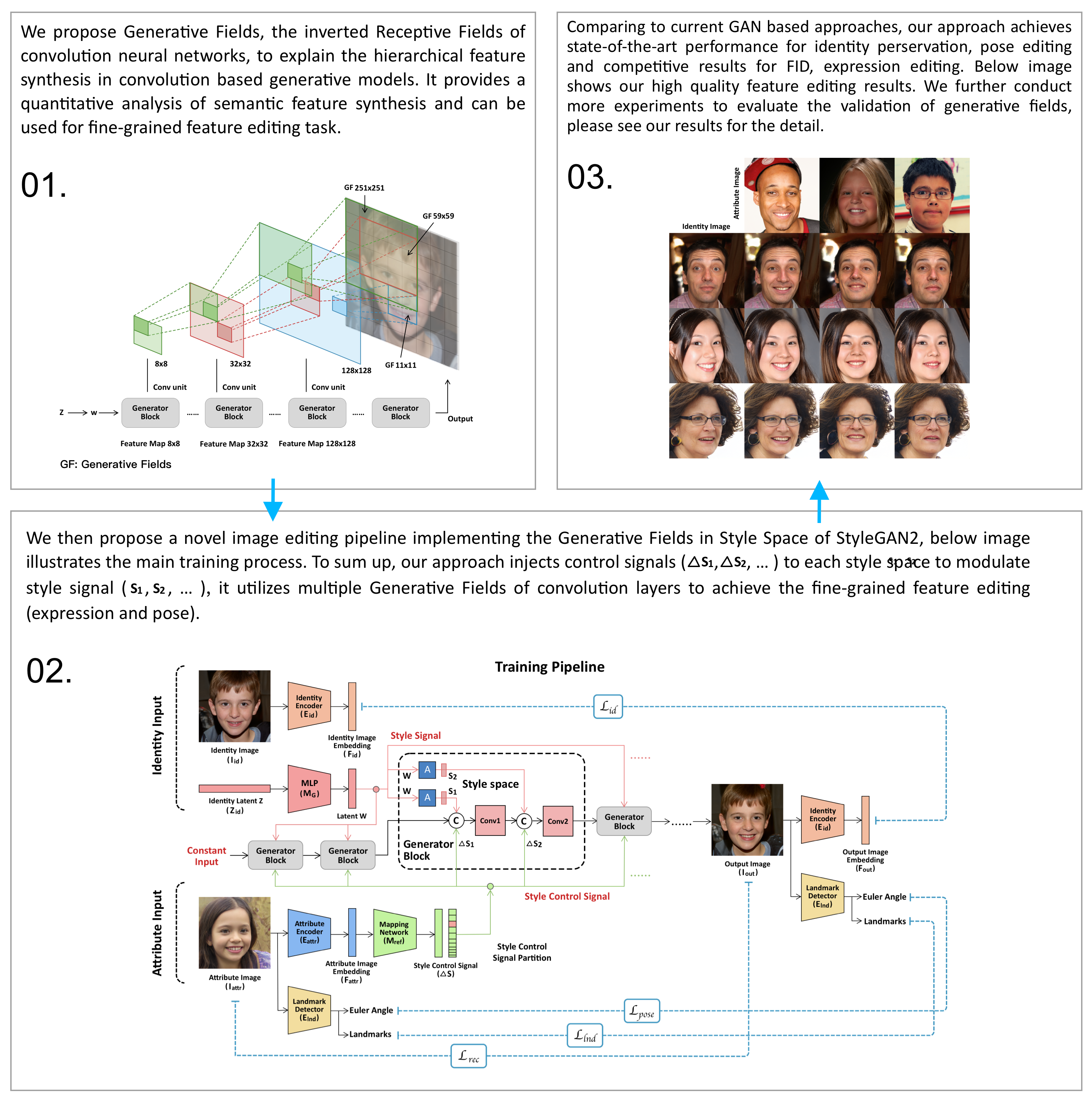}

\end{graphicalabstract}

\begin{highlights}
\item We propose new ``generative fields" inspired by receptive fields of convolution neural networks, to explain the hierarchical feature control with different fineness levels for the input of each convolution layer within the generator, and provide a quantitative experimental analysis.
\item We analyze the principle of style space and connect it with the function of generative fields, showing how it enables fine-grained feature control, and design a new image editing pipeline for the pre-trained StyleGAN2 generator that exhibits improved performance on the image editing task.
\item We evaluate the sparse property of feature editing control in style space, revealing the style space can be used to stack the multiple control signals with different fineness levels.
\end{highlights}


\end{frontmatter}



\section{Introduction}
\label{sec:intro}
Generative adversarial networks (GANs) \citep{goodfellowGenerativeAdversarialNets2014} have seen enormous improvements over the last decade. Approaches such as StyleGAN \citep{karrasStyleBasedGeneratorArchitecture2019a} have demonstrated that it is possible to generate realistic and vivid faces of people who do not exist. One aspect that remains less understood is how to control the fine-grained properties of the generated face, such as head pose or facial expression. A possible solution to this problem is to separate the image generation process from the pose and facial expression transformations and targeted controlling the feature synthesis \citep{karrasStyleBasedGeneratorArchitecture2019a, nitzanFaceIdentityDisentanglement2020a}. However, disentangling the synthesis of content and feature is difficult due to they all come from a sampling in low-dimensional latent space. In this paper, we investigate how image properties are build from convolution neural network intrinsically, revealing that internal aspects of the architecture, namely convolution blocks, have an important impact on how image properties emerge, then utilizing it to control the pose and expression synthesis, achieving Simultaneous Generation and Editing (SGE).

Considering the recent popularity of the AI-generated Content (AIGC) industry, people expect the fine-level control of generated content to merge the AI-based generative pipeline into the existing industrial content-producing pipeline. However, the principles of such controlability are still not well studied, contrasting with standard convolution models. The nature of generative models requires random sampling to provide uncertainty to synthesize new images \citep{kingmaAutoEncodingVariationalBayes2013, goodfellowGenerativeAdversarialNets2014, vandenoordConditionalImageGeneration2016, papamakariosNormalizingFlowsProbabilistic2021a}, and controlling the sampling in latent space is the most common way to edit generated results. The distribution on the latent space is typically a simple low-dimensional probability density, which is necessarily different to the complicated high-dimensional distribution of generated content. Convolution-based generative models like GANs learn the process to transform low-dimensional random samples to high-dimensional synthesized images with photo-realistic quality. This implies a many-to-one relationship that multiple generated features might come from the same dimension in latent space, leading to the entanglement of image features in latent codes \citep{karrasStyleBasedGeneratorArchitecture2019a, liLearningDisentangledRepresentation2019, arjovskyPrincipledMethodsTraining2017a}. This entanglement makes the semantic control of GAN generations a challenge. Current approaches use the concept of learnt latent sampling trajectory \citep{chenInfoGANInterpretableRepresentation2016a, rameshSpectralRegularizerUnsupervised2018, jahanian*SteerabilityGenerativeAdversarial2019}, that regresses a path in low-dimensional space and sample on it to steer the final feature synthesis, or reference-based method, such as a text prompt \citep{liControllableTexttoImageGeneration2019a, zhangStackGANTextPhotoRealistic2017a, 
xiaTediGANTextGuidedDiverse2021a} or an example image \citep{nitzanFaceIdentityDisentanglement2020a, 
dengDisentangledControllableFace2020}, that embeds the reference signal to the low-dimensional latent space, modulating the sampling to edit the image generation. These studies have achieved impressive success in feature editing and style mixing tasks. However, learning a trajectory in latent space is difficult due to the entanglement; a text prompt is a coarse description that cannot provide sufficiently fine-grained information for feature control; and approaches using example images often lead to large deviations from the original synthesised image which requires additional adversarial loss for rectifying.

We get the inspiration from recent investigations into the principle of GANs and specifically StyleGAN \citep{bauGANDissectionVisualizing2018a, wuStyleSpaceAnalysisDisentangled2021, 
heEigenGANLayerWiseEigenLearning2021a}, which uncover the interpretable function of the convolution process and its resulting effect. We extend the concept of receptive fields from convolution neural networks to deep generative models, and use those ``generative fields" to explain the synthesis of visual features at multiple scales quantitatively \citep{karrasStyleBasedGeneratorArchitecture2019a, wuStyleSpaceAnalysisDisentangled2021}. Basing on it, We then propose a new image editing approach that directly adjusts channel-wise style space for a pre-trained StyleGAN2 generator \citep{karrasAnalyzingImprovingImage2020a}, modulating features for two fineness levels, head pose and facial expression, of the generated images by using generative fields. To the best of our knowledge, this is the first work that quantitatively analyzes the role of inverted receptive fields for the feature control of the deep generative model. The main contributions of this work can be summarized as follows:

\begin{enumerate}
    \item We use the concept of ``generative fields", inspired by receptive fields of convolution neural network to explain the hierarchical feature control with different fineness levels for the input of each convolution layer within the generator and provide a quantitative experimental analysis.
    \item We analyze the principle of style space and connect it with the function of generative fields, showing how it enables fine-grained feature control. Then we design a new image editing pipeline for the pre-trained StyleGAN2 generator by using the style space and demonstrate improved performance on the image editing task.
    \item We further evaluate the sparse property of feature editing control in style space, revealing that style space can be used to stack multiple control signals with different fineness levels, in a way consistent with our generative fields analysis.
\end{enumerate}

\section{Related Work}
\subsection{Generative adversarial networks}
Generative Adversarial Networks were proposed in 2014 by \citet{goodfellowGenerativeAdversarialNets2014}, fast becoming one of the important generative models. They are inspired by game theory: two models in which one is generator in charge of generating the result sampling on a probabilistic distribution, the other one discriminator is a critic model in charge of examining the generated result through comparing to the input example; the two models are competing with each other during the training process, making them simultaneously stronger. In 2016, \citet{radfordUnsupervisedRepresentationLearning2015} proposed DCGANs importing the convolution operation into the GANs model, which decreases the computation complexity to synthesize high-resolution images. \citet{karrasStyleBasedGeneratorArchitecture2019a} proposed StyleGAN, a redesign of the generator's structure replacing the initial input with inputs at multiple levels of the generator network, introducing the style signal that is merged into the intermediate generating feature maps by using AdaIN \citep{huangArbitraryStyleTransfer2017} within each generator block. \citet{karrasStyleBasedGeneratorArchitecture2019a} claimed this design can separate content and style generation processes. The quality of StyleGAN was further improved by subsequent versions \citep{karrasAnalyzingImprovingImage2020a, karrasTrainingGenerativeAdversarial2020a} leading to better image quality and training efficiency. Our model leverages the separation of content and style generation in StyleGAN2 \citep{karrasTrainingGenerativeAdversarial2020a} and focuses on investigating fine-grained control of these. We use the pre-trained StyleGAN2 as the backbone and design the auxiliary networks for the control task, which adaptively modulate the style signal while maintaining the high quality of image synthesis.

\subsection{Semantic feature learning and editing}
GAN-based approaches can generate high-fidelity images, but lack the ability to control synthesised features, largely due to the black-box nature of convolution neural networks. In an early study, \citet{zeilerVisualizingUnderstandingConvolutional2014} showed that intermediate layers of convolution neural networks are tuned to image features of various scales and complexity: earlier layers respond to lines, curves, corners and edge/colour conjunction; middle layers respond to more complex textures; latter layers respond to recognizable outlines objects. This showed that deep convolution networks learn structured semantics from training data and interpret images layer by layer gradually from simple lines to complex shapes. \citet{bauGANDissectionVisualizing2018a} further investigates this visualization method for GANs through a relationship-detecting method which can find the relevant components from the convolution feature units and objects in the generated image, demonstrating similar functions in the inverted convolution process. This feature learning process is intuitive for human perception and the learnt features gain the well-defined semantic meaning automatically.

Based on these studies, other works investigate feature editing through various methods. \citet{chenInfoGANInterpretableRepresentation2016a, rameshSpectralRegularizerUnsupervised2018, jahanian*SteerabilityGenerativeAdversarial2019} attempt to use the sophisticated design of regularization for the latent space sampling to disentangle the control of fine-grained feature. \citet{heEigenGANLayerWiseEigenLearning2021a} proposes a sampling vector decomposition method that utilizes the hierarchical feature control of intermediate layers along the forward path, providing an executable method of disentangled feature editing while demonstrating the interpretation property of the convolution-based generative model. \citet{karrasStyleBasedGeneratorArchitecture2019a} analyzes feature decomposition property of the intermediate $\mathcal{W}$ space in StyleGAN, revealing its structural control ability. \citet{nitzanFaceIdentityDisentanglement2020a} based on their research, designing a feature editing pipeline that maps the control vector sampled from the reference image to the intermediate $\mathcal{W}$ space of StyleGAN, modulating the features in final synthesized images. \citet{wuStyleSpaceAnalysisDisentangled2021} investigates the channel-wise input of convolution layers within StyleGAN2, proposing a channel-wise latent style space $\mathcal{S}$ concatenating whole channel inputs among StyleGAN2 generator, revealing the fine-grained feature editing approach of style space. Our model refers to Nizan's main structure \citep{nitzanFaceIdentityDisentanglement2020a} but uses style space $\mathcal{S}$ instead of $\mathcal{W}$ space \citep{wuStyleSpaceAnalysisDisentangled2021} which utilizes generative fields for feature control directly, achieving higher feature decomposition performance, our design will not disrupt the main content generation process, which allows us to remove the adversarial learning process within Nitzan's approach, leading to Simultaneous Generation and Editing (SGE).

\subsection{Receptive fields of convolution neural networks}
Receptive fields in convolution networks measure the extent to which the input signal may affect the output \citep{araujoComputingReceptiveFields2019}; this is inspired by the neuroscience notion from the human visual system. \citet{leWhatAreReceptive2017} propose the calculation method of receptive fields, which is a cumulative process from input image to output feature map, it can measure the receptive scale for output features. Receptive fields are important in designing the structure of convolution networks; for example in object detection models, the object scale should match the appropriate receptive field size for correct feature extraction \citep{coatesSelectingReceptiveFields2011, caoPracticalTheoryDesigning2015}. State-of-the-art approaches such as SSD, YOLO, Faster R-CNN, etc., use anchor-based techniques for multi-scale object detection, the position where the anchor layer should be placed is a critical consideration for the model design due to the influence of receptive fields \citep{girshickFastRCNN2015a, zhangS3FDSingleShot2017a, zhangFaceBoxesCPURealtime2017, redmonYOLOv3IncrementalImprovement2018a}. For the generative model, \citet{jaipuriaRoleReceptiveField2020} analyzed the receptive field function of the GAN's discriminator and its effect on image generation, which is similar to normal convolution networks. In our work, we quantitatively evaluate the influence between inverted receptive fields and feature editing results, disclosing the high relevance between the injected control signal position and the fineness level control in synthesized results.

\section{Method}
\label{sec:method}
\subsection{StyleGAN image generation}
\label{subsec:stylegan}
StyleGAN introduced a hierarchical generative process for generating human faces. Contrary to classical GAN approaches the generating noise is injected at multiple processing blocks separately. In this paper, we use the version proposed in StyleGAN2 \citep{karrasTrainingGenerativeAdversarial2020a}, although the approach could be used similarly on the original StyleGAN.

\begin{figure}[htb]
\centering
    \includegraphics[width=0.7\textwidth]{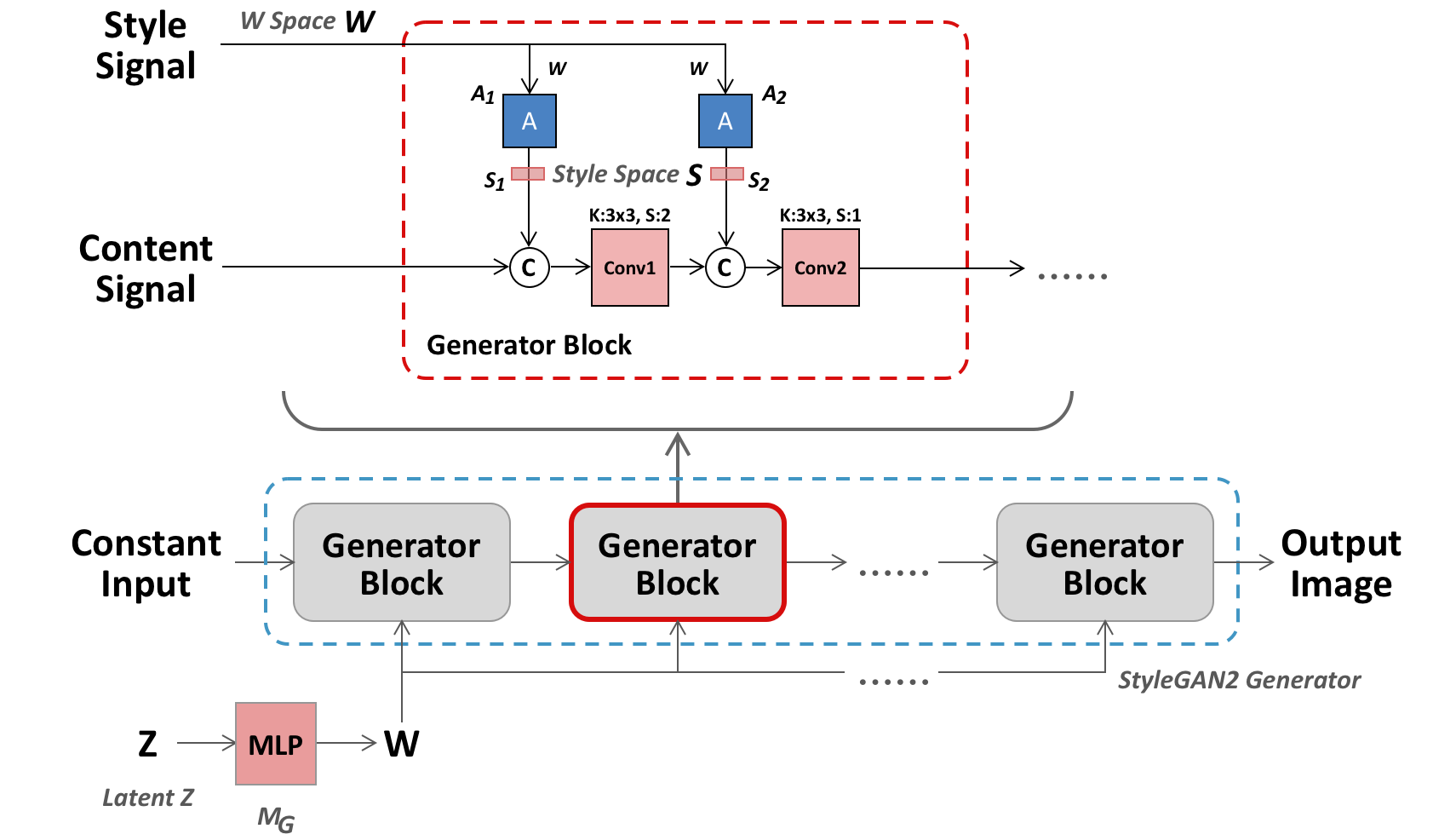}
    \caption{Image generation process in StyleGAN2. Bottom shows the whole generation pipeline including content process and style sampling process; top shows the detailed style modulation process.}
    \label{fig:fig01}
\end{figure}

The StyleGAN2 architecture, illustrated in Figure~\ref{fig:fig01}, is composed of several generator blocks~\citep{karrasAnalyzingImprovingImage2020a}. In contrast to typical GANs, the \textit{content signal} input to the first block is a constant vector only responsible for generating rough output; the main factor driving the variety and vividness of generated images is the \textit{style signal}. Specifically, each generator block except the first consists of two affine transformations $A_1, A_2$ and two convolution layers \texttt{conv1}, \texttt{conv2} with same the kernel size $3\times3$ and different strides ($2, 1$ respectively); the first generator block only has one convolution layer with $3\times3$ kernel size and stride of $1$; we omit the RGB convolution layer in each block since they are not relevant of controlling the main feature synthesis~\citep{wuStyleSpaceAnalysisDisentangled2021}. Each generator block upscales the input feature map to twice its size, the number of generator blocks $N$ is therefore dependent on the desired resolution of the synthesized image. We use the $256\times256$ resolution model composed of $N=7$ generator blocks (13 convolution layers for style space~\citep{wuStyleSpaceAnalysisDisentangled2021}) in our experiments, we denote the convolution layers as \texttt{conv0} to \texttt{conv12}. For a pre-trained StyleGAN2 model, the image generation process is that the content signal is set from the initial learnt constant and gradually builds the feature map through each generator block; the style signal, a 512-dimensional vector $z$, is sampled randomly from initial latent space $\mathcal{Z}$ and then transformed by a dense model $M_G$ to gain the latent vector $w$ for better feature disentanglement \citep{karrasStyleBasedGeneratorArchitecture2019a}, forming a new latent space $\mathcal{W}$. The latent vector $w$ is replicated for the number of generator blocks to be the input of each one, providing the variety of the style signal, the overall latent vectors $w$ form another latent space $\mathcal{W}+$ \citep{wuStyleSpaceAnalysisDisentangled2021}, we denote it as $W_+$.

Within each generator block, for each convolution layer \texttt{conv\textit{i}} (\textit{i} $\in [1, 2]$), an affine transformation $A_{i}$ is learnt that maps the input latent vector $\mathbf{w}$ to a style vectors $\mathbf{s}_i$ with dimensions equal to the input channel of \texttt{conv\textit{i}}~\citep{karrasAnalyzingImprovingImage2020a}. Each style vector provides functional style information to modulate the visual feature synthesis\footnote{The specific style modulation operation varies for different StyleGAN versions: StyleGAN1 uses AdaIN \citep{huangArbitraryStyleTransfer2017}, whereas StyleGAN2 uses mod-demod operations \citep{karrasAnalyzingImprovingImage2020a}. We omit the noise signal modulation since it doesn't control the main factor of synthesized features.} \citep{wuStyleSpaceAnalysisDisentangled2021}. All style vectors form a new latent space $\mathcal{S}$, of dimension 4928 for our architecture.\footnote{We use the definition of style space from \citet{wuStyleSpaceAnalysisDisentangled2021} where the dimension is corresponding to all input channels of \texttt{conv1}, \texttt{conv2} from each generator block.} To simplify notation, we define the overall 4928-dimensional style vector $S=[S_d]_{1\leq d \leq 4928}$ as the concatenation of the style vectors for all convolution layers and $A$ as the complete mapping from $W_+$ to $S$ such that $S = A( W_+ )$.
\label{term:stylesignal}

\subsection{Generative fields}
\label{subsec:generativefields}
Whether applied to classification or generation tasks, training convolution units implies learning spatial semantic information. The locality of convolution kernels implies that features at different layers encode patterns of different granularity. For generative models, early layers control global features whereas later layers control more local features (the converse is true for classification CNNs) \citep{zeilerVisualizingUnderstandingConvolutional2014, bauGANDissectionVisualizing2018a, karrasStyleBasedGeneratorArchitecture2019a, wuStyleSpaceAnalysisDisentangled2021, heEigenGANLayerWiseEigenLearning2021a}. 
In the case of perceptive CNNs, this is explained by the concept of receptive fields \citep{yuMULTISCALECONTEXTAGGREGATION2016a}.
We propose to extend the notion of receptive fields to generative models, coined \textit{generative fields} in this work. Namely, the feature generation in convolution-based generative models is influenced by the scale of receptive field in the output, which is inverted compared to, eg. object detection models. Equation \ref{eqn:eqn01} is the formula definition of generative field size inspired from receptive field calculations~\citep{araujoComputingReceptiveFields2019}:

\begin{equation}
\begin{gathered}
    g_{0}=\sum_{l=1}^{N-L} \left( (k_{N-l+1}-1)\prod_{i={N-l+1}}^{N}s_i \right)+1
\end{gathered}
\label{eqn:eqn01}
\end{equation}
where $k_l, s_i$ are the kernel and stride size from $l$-th, $i$-th layer of convolution calculation , $g_{0}$ is the generative field size of input feature map of $(L+1)$-th convolution layer (L $\in [0, 12]$), $N$ is the total number of  convolution layers (we use the start point of $L$ from 0th for the consistency of style space).
\vspace{0.5em}

\begin{figure}[htb]
\centering
    \includegraphics[width=0.8\textwidth]{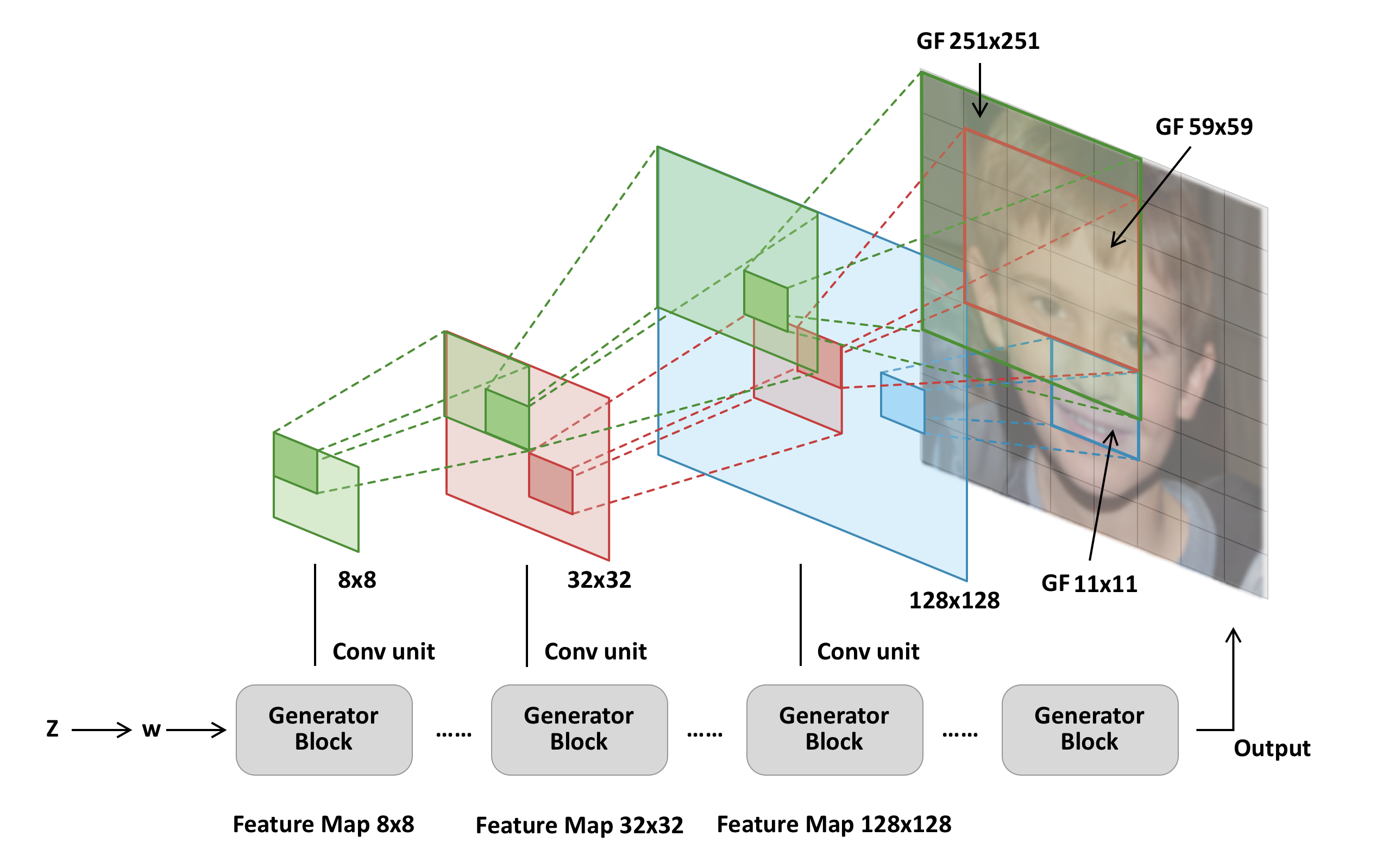}
    \caption{Generative fields produced by convolution units at different StyleGAN2 generator blocks. The leftmost unit feature map size is $8\times8$ and controls the largest generative field, of size $251\times251$; conversely, the rightmost unit feature map size is $128\times128$ and controls the smallest generative field, of size $11\times11$.}
    \label{fig:fig02}
\end{figure}

As the example of feature generation in StyleGAN2, the position of convolution unit in the generator network determines various generative fields. Figure~\ref{fig:fig02} depicts the generative field of chosen convolution layers for input feature map of sizes $8\times8, 32\times32, 128\times128$ in the StyleGAN2 generator. As illustrated in section \ref{subsec:stylegan}, all convolution layers can be indexed from \texttt{conv0} to \texttt{conv12} for $256\times256$ resolution model, we calculate generative fields of convolution layers whose indices are \texttt{conv2}, \texttt{conv6}, \texttt{conv10}, the \nth{3}, \nth{7}, \nth{11} convolution layer of StyleGAN2 generator, giving generative field sizes of $251\times251, 59\times59, 11\times11$ respectively, which are the typical size corresponding to large, middle, small features in a model generating images of resolution $256\times256$. The content generation process of StyleGAN2 is a composition of synthesized features with generative fields sizes from small to large, combining all information in a coherent whole. The detailed generative field size for whole convolution units is provided in the appendix.

Specifically, a convolution layer with large generative field size (eg, \texttt{conv2}) should allow the control of global style features, such as head pose, because its generative field size covers the whole image. Similarly, convolution layers with smaller generative fields (eg, \texttt{conv6}) should control more fine-grained style features such as facial expression. Finally, convolution layers with the smallest generative fields (eg, \texttt{conv10}) should have no structural impact, but refine the tone and texture of the generated image. Intuitively, this analysis is compatible with the style mixing experiments in the original StyleGAN paper~\citep{karrasStyleBasedGeneratorArchitecture2019a}. On the other hand, style vectors as the input of $\texttt{conv2}, \texttt{conv6}, \texttt{conv10}$ in style modulation process determine the editing information with various fineness level for final result, indicating that the style space $\mathcal{S}$ formed by all style vectors (Sec.~\ref{term:stylesignal}) could be used to edit synthesized feature in different fineness level.


\newcommand{\SG}{\ensuremath{G}}
\newcommand{\Ei}{\ensuremath{E_\textnormal{id}}}
\newcommand{\Ea}{\ensuremath{E_\textnormal{attr}}}
\newcommand{\Mr}{\ensuremath{M_\textnormal{ref}}}
\newcommand{\El}{\ensuremath{E_\textnormal{lnd}}}
\newcommand{\Ii}{\ensuremath{I_\textnormal{id}}}
\newcommand{\Ia}{\ensuremath{I_\textnormal{attr}}}
\newcommand{\Io}{\ensuremath{I_\textnormal{out}}}
\newcommand{\Fi}{\ensuremath{F_\textnormal{id}}}
\newcommand{\Fa}{\ensuremath{F_\textnormal{attr}}}
\newcommand{\Fo}{\ensuremath{F_\textnormal{out}}}
\newcommand{\ds}{\ensuremath{\Delta S}}
\newcommand{\Zi}{\ensuremath{Z_\textnormal{id}}}

\subsection{Feature control for StyleGAN2}
\begin{figure}[htb]
\centering
    \includegraphics[width=0.6\textwidth]{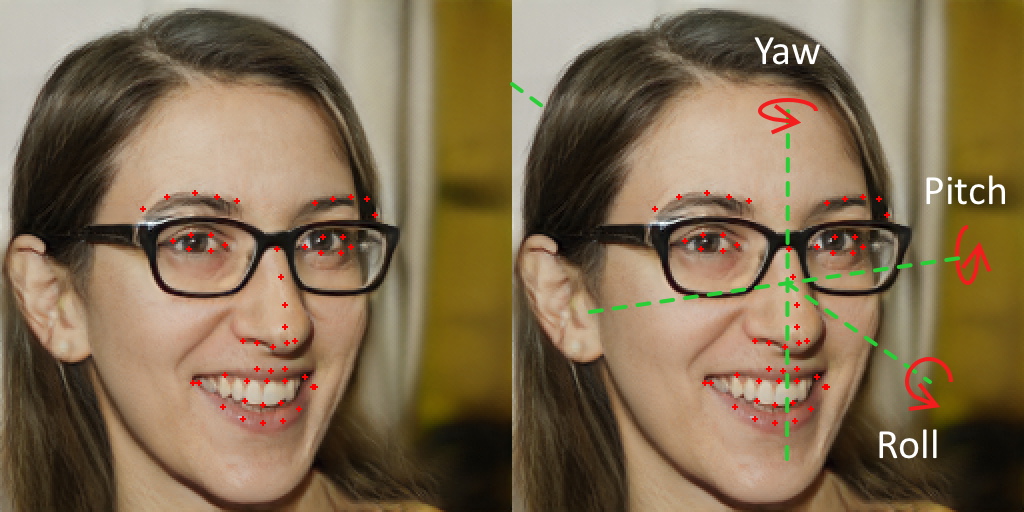}
    \caption{Facial landmarks (left) and head pose Euler angles (right).}
    \label{fig:fig03}
\end{figure}

Based on our analysis of generative fields (Sec.~\ref{subsec:generativefields}), we design a head feature editing pipeline working on style space $\mathcal{S}$ of a pre-trained StyleGAN2, adaptively utilizing channel-wise latent space with multiple generative fields for fine-grained control of human face synthesis. Given a reference image as the style input, head pose and facial expression features are extracted by a set of neural networks as the control signal, which is then injected into the style space of each convolution generator block, adjusting the style modulation during generating. Another image coupled with its sample vector provides identity information for the synthesis process. In order to encode facial expression and head pose features, we use a combination of facial landmarks and head pose Euler angles, as illustrated in Figure~\ref{fig:fig03}.

The facial landmarks are defined by 68 feature points on the human face, used to encode the facial pose and expression information. Each feature point is denoted by 3-dimensional point coordinates. Due to the first 17 landmarks being dependent on face shape (and therefore identity) rather than expression, we only use 51 inner-landmarks from index 18 to 68 for better editing results, which are shown as red colour points in Figure~\ref{fig:fig03}. The right panel illustrates the definition of Euler angles: yaw, pitch and roll. The each angle is in the range $(\text{-}\frac{\pi}{2}, \text{+}\frac{\pi}{2})$. These angles are used to define the head pose, combined with facial landmarks to describe the overall human facial features.

\subsubsection{Feature control architecture}
\label{subsec:architecture}
\begin{figure*}[htb]
\centering
    \includegraphics[width=\textwidth]{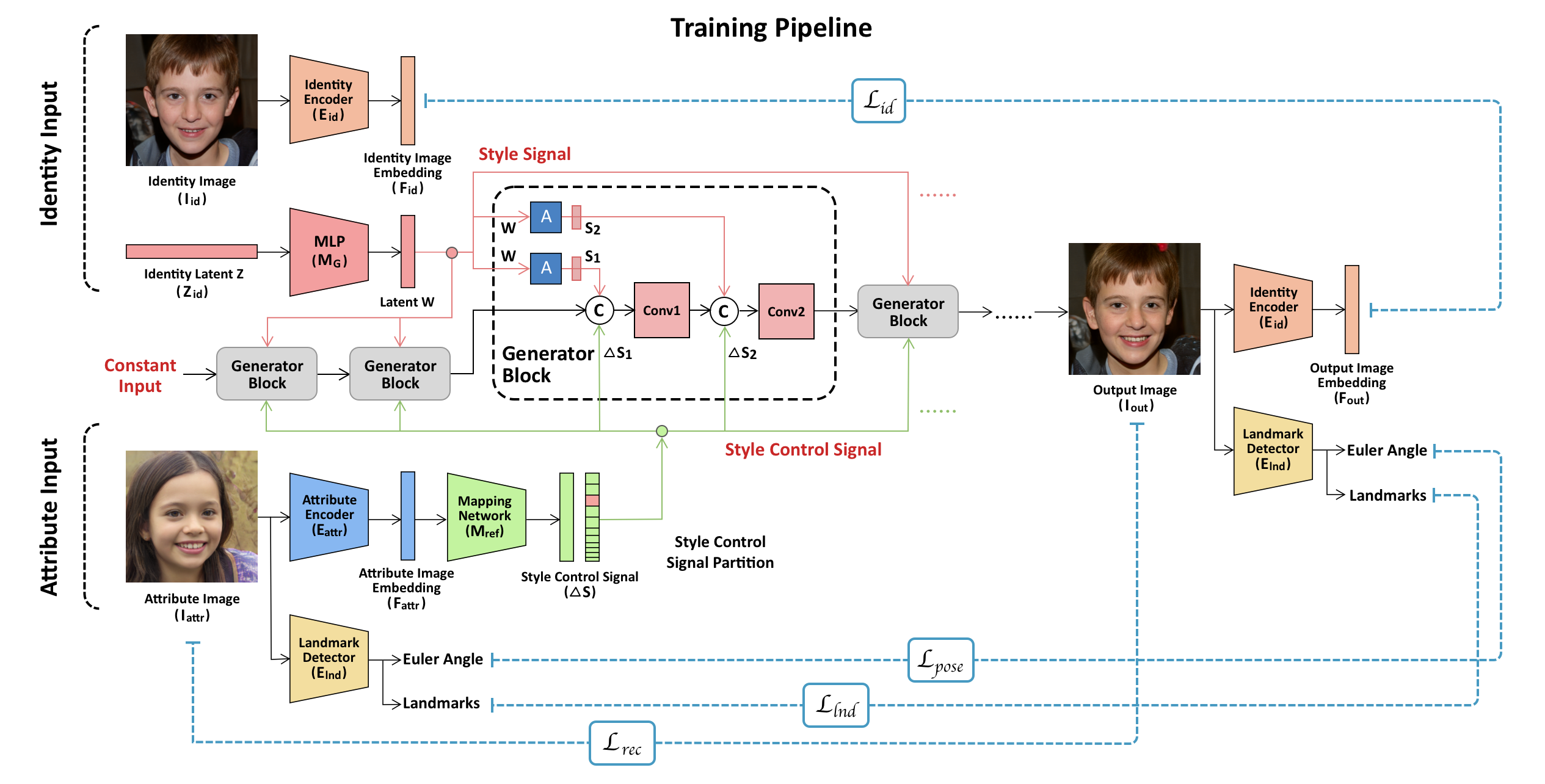}
    \caption{Image editing pipeline for StyleGAN2 using style space $\mathcal{S}$. Identity input including latent vector $Z$ and corresponding generated image $I_{id}$ for the facial generation with identical features. The attribute input is a reference image $I_{attr}$ from which we extract facial features (expression, head pose) for controlling the image generation. All control signals work within each generator block, modulating the style signal samples in layer-wise style space $\mathcal{S}$.}
    \label{fig:fig04}
\end{figure*}

Our image editing pipeline, shown in Figure~\ref{fig:fig04}, extends the approach proposed from \citet{nitzanFaceIdentityDisentanglement2020a} by 1) controlling explicitly for pose, and 2) modulating the control signal at each generator block. We modify the StyleGAN2 generator \citep{karrasAnalyzingImprovingImage2020a} to input the \textit{control signal}, containing face pose and expression, into the style space. 

The architecture is composed of 5 networks: 
The StyleGAN2 generator \SG, 
the identity encoder \Ei, 
the attribute encoder \Ea, 
the reference mapping network \Mr, 
and the facial landmark detector \El. 
In our approach, \SG, \Ei, \Ea and \El~are pre-trained and only \Mr~is trained from scratch. 
There are 2 types of input, 1) identity input consists of the StyleGAN-generated image $\Ii=\SG(A(M_{G} (\Zi)))$ that we want to edit and its latent sampling code \Zi~for StyleGAN2; 2) the attribute input \Ia, which is a reference image providing the style information that the identity image should be changed to mimic. 
During training, the networks \Ei{} and \Ea~will encode the images \Ii{} and \Ia~into the feature representations \Fi and \Fa, respectively:
\begin{eqnarray}
    \Fi & = & \Ei(\Ii) \\
    \Fa & = & \Ea(\Ia)
\label{eqn:eqn02}
\end{eqnarray}

\label{term:stylecontrolsignal}
\Mr~learns a transformation from the feature embedding of the attribute image \Fa~to the style space $\mathcal{S}$, to produce a control signal that modulates the style signal \Ii, editing the features by offsetting the channel-wise numerical value \citep{wuStyleSpaceAnalysisDisentangled2021}. The StyleGAN2 generator $G$ then synthesizes the edited image $I_{out}$ through modulated style vectors: 
\begin{eqnarray}
    \ds & = & \Mr(\Fa) \\
    S_{id} & = & A( M_{G} (\Zi) ) \\
    \Io & = & \SG( S_{id} + \ds ) 
\label{eqn:eqn03}
\end{eqnarray}
where \Zi~is used to recover the generation of identity image in StyleGAN2, $S_{id}$ is the style signal gotten from \Zi, $M_{G}$ and $A$ denote the latent mapping network of StyleGAN2 and affine transformations of generator blocks. Referring to style signal definition (Sec.~\ref{term:stylesignal}), we define $\Delta S = [\Delta s_d]_{1\leq d \leq 4928}$ as the style control signal concatenating all style control vectors $\Delta \mathbf{s}$ across all generator blocks, which is 4928-dimensional vector corresponding to the dimension of $S_{id}$.

In contrast, \citet{nitzanFaceIdentityDisentanglement2020a} only uses $\mathcal{W}$ space as the controlling vector space and therefore requires an additional adversarial learning process to regularize the initial sampling, whereas our method uses the style space that just provides the control signal rather than the whole style signal, dividing up the image synthesizing and editing completely, achieving the ability of simultaneously generating and editing (SGE) during inference.

\subsubsection{Loss functions}
The loss functions we used are modified from \citet{nitzanFaceIdentityDisentanglement2020a} to account for the differences in our architecture. Considering the head pose rotation is a movement in 3D space, we use a 3D landmark detector from \citet{zhuFaceAlignmentFull2019} to extract 3D facial landmarks and Euler angles from facial images. Moreover, because Nitzan's approach predicts the new $\mathcal{W}$ space vector conditioned on the attribute image input, the new $\mathcal{W}$ vector can deviate from the original data manifold failing to produce a valid face for the following task. They resolve this by pre-training the mapping network with an adversarial loss \citep{nitzanFaceIdentityDisentanglement2020a}. One benefit our editing features in style space directly is that our approach preserves the content synthesis signal, and therefore does not require the addition of an adversarial loss.

\vspace{0.5em}
\noindent \textbf{Identity loss:} \hspace{1pt} $\mathcal{L}_{id}$, is used to ensure that the transformation performed by the network preserves the identity of the person in the input image. It is implemented as the $L_1$ distance between the identity embedding of the identity image \Fi=\Ei(\Ii) and of the generated image \Fo=\Ei(\Io):
\begin{equation}
    \mathcal{L}_{id}=\,\parallel E_{id}(I_{id})-E_{id}(I_{out}) \parallel_1
    \label{eqn:eqn04}
\end{equation}

\noindent \textbf{Attribute loss:} \hspace{1pt} $\mathcal{L}_{attr}$ is used to ensure that the generated image's pose and expression are similar to the attribute image's. It is implemented as the combination of the $L_2$ distance between detected facial landmarks and estimated Euler angles $\alpha, \beta, \gamma$ between \Ia{} and \Io. 

\begin{equation}
\begin{gathered}
    \mathcal{L}_{lnd}=\,\parallel E_{lnd}(I_{attr})-E_{lnd}(I_{out}) \parallel_2\; \\
    \mathcal{L}_{pose}=\,\parallel E_{pose}(I_{attr})-E_{pose}(I_{out}) \parallel_2 \\
    \mathcal{L}_{attr}=\mathcal{L}_{lnd}+\mathcal{L}_{pose}, 
\end{gathered}
\label{eqn:eqn05}
\end{equation}
where $E_{lnd}, E_{pose}$ are all provided by pre-trained 3D landmark and pose detectors.

\vspace{0.5em}
\noindent \textbf{Reconstruction loss:} \hspace{1pt} We use the reconstruction loss $\mathcal{L}_{rec}$ from \citet{nitzanFaceIdentityDisentanglement2020a} to preserve pixel-level information if $I_{id}, I_{attr}$ are same, which is a weighted sum of pixel-wise $L_{1}$ loss and $\mathrm{MS\text{-}SSIM}$ loss: 

\begin{equation}
\begin{gathered}
\mathcal{L}_{struc}=\mathrm{MS\text{-}SSIM}(I_{attr}, I_{out})) \\
\mathcal{L}_{mix}=\alpha \mathcal{L}_{struc} + (1-\alpha)\parallel I_{attr} - I_{out} \parallel_1
\end{gathered}
\label{eqn:eqn06}
\end{equation}
where $\alpha$ is the mixing hyperparameter, for which we use 0.84 as suggested by \citet{zhaoLossFunctionsImage2017}.
\vspace{0.5em}

The training process is divided into the cross stage and the reconstruction stage, where the former is training for pose and expression transfer and the latter is training for content consistency. During cross stage the images \Ii{} and \Ia{} are different and the model is trained to generate the image capturing features from \Ia; during the reconstruction stage the images \Ii{} and \Ia{} are the same and the model is trained to reconstruct the inputs, $L_{mix}$ is only employed when \Ia = \Ii:

\begin{equation}
\mathcal{L}_{rec}=\left\{
\begin{aligned}
    &\mathcal{L}_\textnormal{mix}, & \Ia=\Ii \\
    &0, &\textnormal{otherwise}
\end{aligned}
\label{eqn:eqn07}
\right.
\end{equation}
\vspace{0.1em}

\noindent \textbf{Total loss} \hspace{1pt} The overall loss is the weighted sum of the above losses, 3 parameters $\lambda_1$, $\lambda_2$, $\lambda_3$ are used to control the factor of each one, influencing the identity preservation, attribute feature editing and general attribute preservation in generated results:
\begin{equation}
\mathcal{L}_\textnormal{total}=\lambda_1\mathcal{L}_\textnormal{id}+\lambda_2\mathcal{L}_\textnormal{attr}+\lambda_3\mathcal{L}_\textnormal{rec}
\label{eqn:eqn08}
\end{equation}

\subsection{Style space regularization}
We found that training could become unstable after several epochs. We explain it through the derivation of data manifold in style space. The output image should capture the facial landmarks of \Ia, however, the loss produced from a bad-quality output image could be lower than a high-quality output image because \El{} may also recognize a bad-quality human face and detect its landmarks, leading to the modulated value of style space deviating from the data manifold.

To fix it, we propose a style regularization term which models the sampling of each style space channel as a Gaussian distribution (and for simplicity, assumed independent). Because the sampling value of the style space for the training dataset is known, we can get the mean $\mu_i$ and standard derivation $\sigma_i$ of the training examples for each style channel $s_i$. We therefore can also compute the log-likelihood $L(S)$ for each style space channel $s_i$ on every training batch, given by
\begin{equation}
    L(S)=-\frac{1}{2\sigma_i^2}\sum_{i}^{|S|}(s_i-\mu_i)^2
    \label{eqn:eqn09}
\end{equation}
where $\mu_i, \sigma_i$ are mean and standard derivation of style code from the generated dataset, $s_i$ is the style sampling of $i$-th channel of style space.
\vspace{0.5em}

We can then maximise the log-likelihood to restrict the drift from the data manifold when searching style vectors. The style regularization term $-L(S)$ is added to the loss function during optimization. In our ablation experiments, we show this noticeably improves editing performance.

\subsection{Implementation details}
We use a pre-trained StyleGAN2 model with $256\times256$ resolution in all experiment tasks, the training pipeline is composed of 2 stages, cross stage and reconstruction stage, and a hyper-parameter controls the ratio of them in charge of the attribute learning and reconstruction respectively, we choose the ratio 3 as the suggestion of referred work \citep{nitzanFaceIdentityDisentanglement2020a}. A pre-trained landmark detection model 3DDFA is used as $E_{lnd}$ \citep{zhuFaceAlignmentFull2019} to regress 3D coordinates of 68 facial key points and 3 Euler angles, facilitating the feature editing and embedding experiment. 

The Adam optimizer \citep{kingmaAdamMethodStochastic2015} is used for all experiments with parameters $\beta_1$=0.9, $\beta_2$=0.999 and the learning rate is dynamically changed from $5e-6$ to $1e-7$ for quick convergence. We empirically set the loss weights to $\lambda_1=1,\lambda_2=0.01,\lambda_3=0.02$ and the hyper parameter of $\mathcal{L}_\textnormal{rec}$ to 0.84. For the calculation of $\mathcal{L}_\textnormal{lnd}$, we take the suggestion from the referred work to use 51 inner-face landmarks \citep{nitzanFaceIdentityDisentanglement2020a} as they report the jawline landmarks could strongly result in the unreal face generation. The model is trained with batch-size 6 on a single NVIDIA GeForce RTX 3090 GPU and the training process is very efficient, taking only one day to converge.

\section{Experiments}
\label{sec:exp}
\subsection{Image editing}
We conducted comprehensive experiments to evaluate our image editing approach and to verify generative fields theory. For the experiment design, a pre-trained StyleGAN2 generator would generate high-quality images with the identity of identity image input whose facial features are controlled by another attribute image input, the output should maintain the identity from identity input, and simultaneously follow the facial features (pose, landmarks) from attribute input.

The metrics for the evaluation consist of two aspects, \textit{realism} and \textit{accuracy}. To measure the realism of generating content, we use Fréchet inception distance (FID) \citep{heuselGANsTrainedTwo2017} for image quality evaluation which compares the distribution divergence between generated data and real-world FFHQ256 dataset: if the generated data is realistic, the FID score should be low. 

The accuracy aspect is composed of identity accuracy, expression accuracy and pose accuracy. 
First, \textit{identity accuracy} is measured as the cosine distance between the embedding representation of the identity and edited images. 
Second, \textit{expression accuracy} is measured as the Euclidean distance between the facial landmark locations for the attribute and edited images. We normalize the landmarks by dividing the image resolution for the consistency of comparison with previous methods \citep{gabbayDemystifyingInterClassDisentanglement2019, nirkinFSGANSubjectAgnostic2019a, nitzanFaceIdentityDisentanglement2020a}. 
Third, \textit{pose accuracy} is measured as the mean squared error deviation between the Euler angles of the attribute and edited images. 

\begin{figure}[htb]
\centering
    \includegraphics[width=0.55\textwidth]{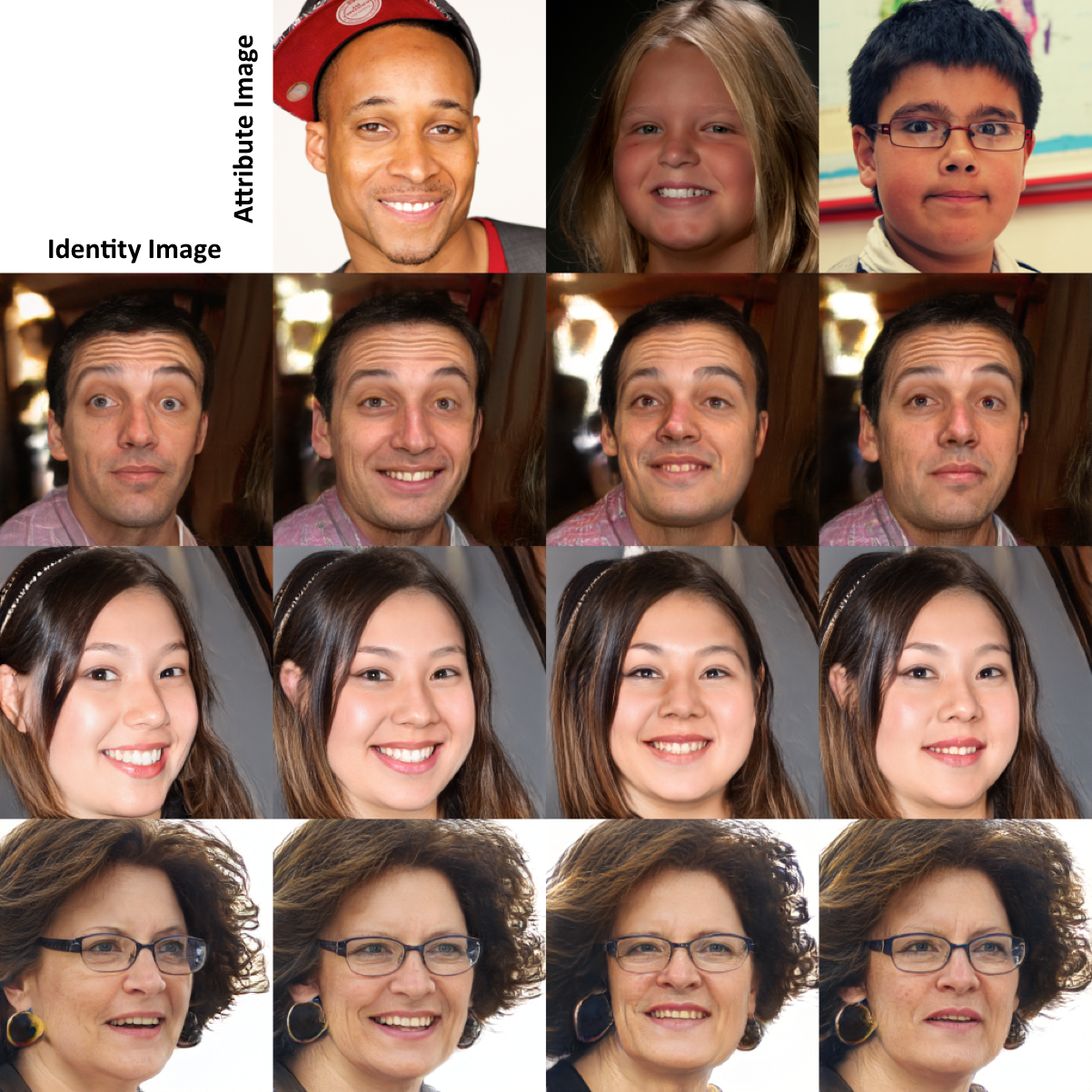}
    \caption{Image editing result. Identity images are generated from StyleGAN2 randomly, attribute images are the real image set sampled from FFHQ256 dataset, identity images should capture pose and expression from attribute images.}
    \label{fig:fig05}
\end{figure}

Considering the diversity of StyleGAN2, we use the generated dataset from pre-trained StyleGAN2 on FFHQ256 dataset, consisting of 48,000 high-quality synthesized images as the training data for our image editing pipeline, from which 36,000 images are used as identity images and 12,000 images as attribute images. This ensures that a wide range of attribute variation is covered by model generation capacity. We use real-world attribute images from the FFHQ256 dataset for evaluation to test the model performance and generalization. Figure~\ref{fig:fig05} demonstrates high-quality image editing results: the generated image captures the pose and expression of attribute images while preserving the identity of identity images.

\subsubsection{Comparison with previous methods}
We compared the proposed approach's generations with previous approaches: LORD \citep{gabbayDemystifyingInterClassDisentanglement2019}, FSGAN \citep{nirkinFSGANSubjectAgnostic2019a} and IDDISEN \citep{nitzanFaceIdentityDisentanglement2020a}. We measured the realism (using FID), identity preservation and expression and pose fidelity. We design an attribute evaluation experiment to measure the model performance for identity preservation, expression editing and pose editing, specifically, we randomly edit identity images referring to attribute images for 1,000 samples on the best overall performance configuration, then calculate average metrics between inputs and edited output, other models follow the similar evaluation approach. As shown in table \ref{tab:tab01}, our approach achieves state-of-the-art performance for identity preservation, pose editing and competitive results for FID, expression editing.
\begin{table}[h]
  \centering
  \resizebox{0.7 \linewidth}{!}{
    \begin{tabular}{l l l l l}
    \toprule
    Method & FID $\downarrow$ & Identity $\uparrow$ & Expression $\downarrow$ & Pose $\downarrow$ \\
    \midrule
    LORD      & 23.08       & 0.20        & 0.085       & 13.34 \\
    FSGAN     & 8.90        & 0.35        & \textbf{0.013} & 6.73 \\
    IDDISEN   & \textbf{4.28} & 0.60        & 0.017       & 9.74 \\
    Ours      & 4.89        & \textbf{0.82} & 0.017       & \textbf{1.67} \\
    \bottomrule
  \end{tabular}
  }
  \caption{Quantitative evaluation with previous methods}
  \label{tab:tab01}
\end{table}
\vspace{-1em}

\subsubsection{Ablation study}
We conducted an ablation study to evaluate the importance of the different components of our image editing pipeline, following the same approach in the last sub-section, and summarise the results in table~\ref{tab:tab02}.
\begin{table}[h]
  \centering
  \resizebox{0.8 \linewidth}{!}{
    \begin{tabular}{l l l l l}
    \toprule
    Configuration & FID $\downarrow$ & Identity $\uparrow$ & Expression $\downarrow$ & Pose $\downarrow$ \\
    \midrule
    default configuration   & 3.65 & 0.84 & 0.026 & 3.29 \\
    -- feature concatenation & 8.52 & 0.77 & 0.016 & 1.62 \\
    + style regularization& 4.48 & 0.83 & 0.022 & 2.33 \\
    + Euler angle loss& 4.89 & 0.82 & 0.017 & 1.67 \\
    \bottomrule
  \end{tabular}
  }
  \caption{Quantitative evaluation of image editing pipeline}
  \label{tab:tab02}
\end{table}

\begin{figure}[h]
\centering
    \includegraphics[width=0.54\textwidth]{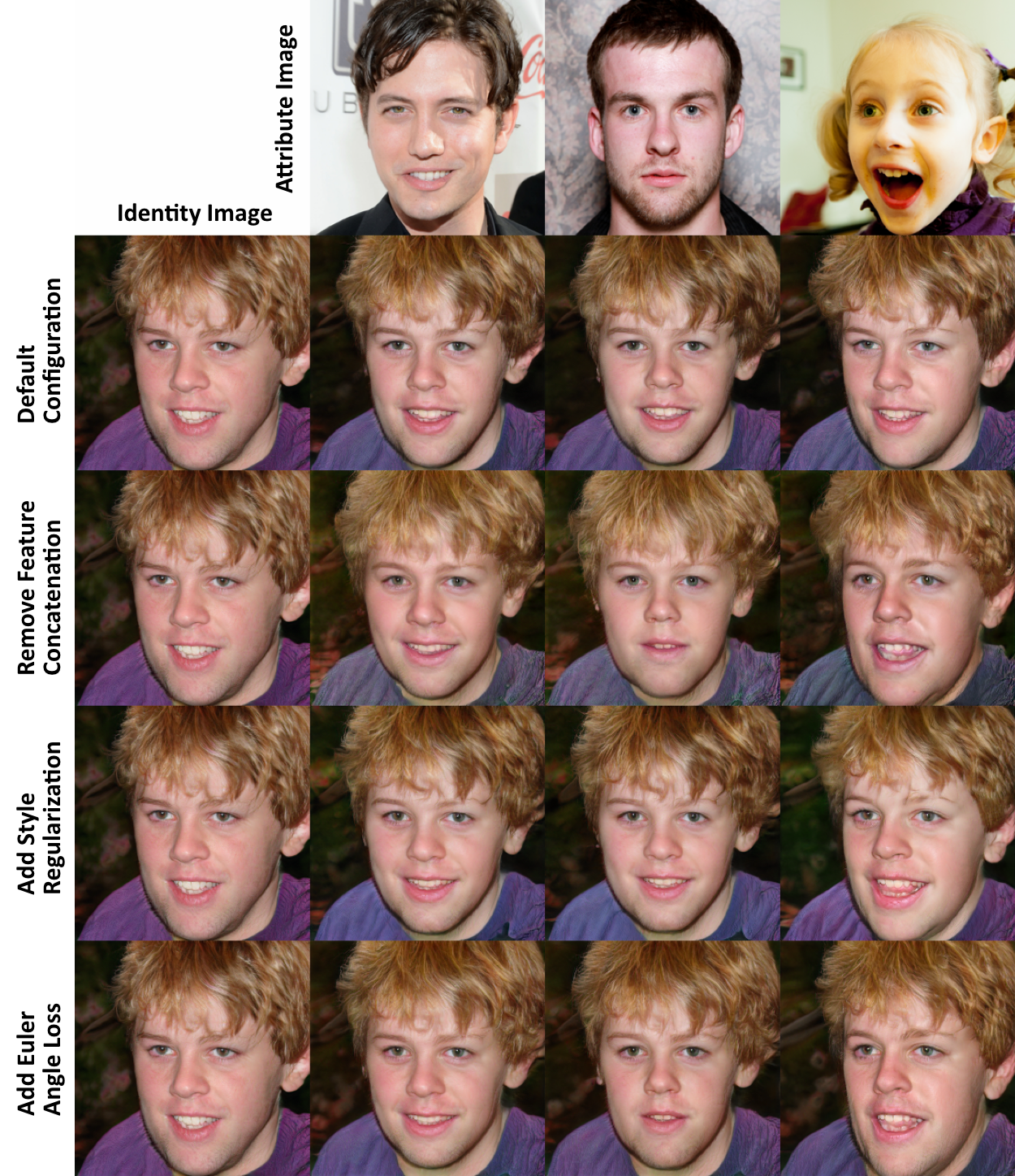}
    \caption{Comparison of image editing results with different experiment configurations in the ablation study. The second row results significantly improve the editing performance by removing the feature concatenation but loses some quality. The last row adds style regularization and Euler angle loss, getting the best overall performance with the balance between image quality and editing.}
    \label{fig:fig06}
\end{figure}

In this table, the default configuration is based on \citet{nitzanFaceIdentityDisentanglement2020a} approach where the input of reference mapping network $M_{ref}$ is the concatenation of the identity image feature $F_{id}$ and attribute image feature $F_{attr}$ and the output is fed to style space $\mathcal{S}$ for image editing (Sec.~\ref{subsec:architecture}). 
The other configurations are adjustments to the default configuration. We found that removing the feature concatenation of two images embedding can significantly improve the editing performance in the experiment, we explain that our new design uses style space to do the feature editing, the generation signal has been provided by recorded $Z_{id}$, which is not necessary to give more identity information for generator, therefore the new scheme completely decouple the content generation and editing. Furthermore, we found the usage of style regularization and Euler angle loss can get the best performance with the balance between the quality and editing. Figure~\ref{fig:fig06} illustrates the visual evaluation of performance improvement, the experiment configuration from top to bottom follows the sequence of table~\ref{tab:tab02}.

\subsection{Generative fields}
\subsubsection{Sparsity of style space control}
The generative fields theory tells us that feature synthesis with various scale and granularity depends on the generative fields of the convolution units, identical content should have same generating process including content and style. As the edited image preserve the most identity of the source image, we could make a hypothesis that the modification of style signal for edited image is small and the style control signal should be sparse. We test this hypothesis with two experiments to evaluate how sparse is the editing signal, and the consistency of active style dimensions for multiple editing tasks.

\begin{figure}[h]
\centering
    \includegraphics[width=0.8\textwidth]{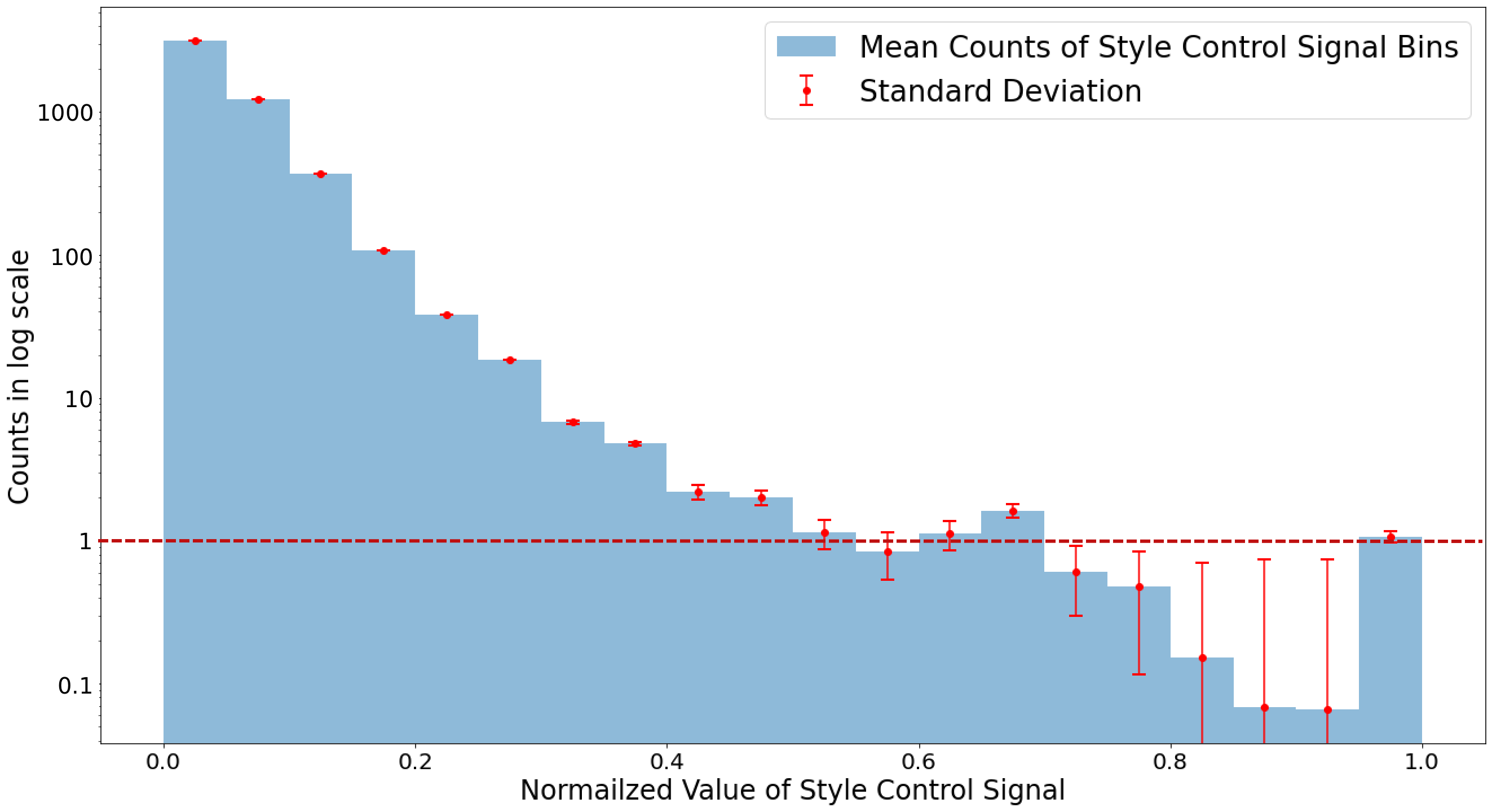}
    \caption{Histogram of mean style control signal $\Delta S$. X-Axis is the normalized value of $\Delta S$, Y-Axis is the mean counts of $\Delta S$ among all experiments locating on 20 data bins.}
    \label{fig:fig07}
\end{figure}

For the first experiment, we calculate the mean histogram of all style control signals $\Delta S$ ($4928 \times 1$ dimensions, Sec.~\ref{term:stylecontrolsignal}) from multiple feature editing tests. Specifically, we conduct the experiment with attribute images $I_{attr}$ from the whole real dataset FFHQ256 and a single identity image $I_{id}$ from our synthesized identity dataset. We first calculate the histogram data of $\Delta S$ value in each test, due to the different range and scale of $\Delta S$ in each experiment we use the absolute value of $\Delta S$ and normalize it for consistency across all tests. Therefore, the higher values indicate larger contributions for the feature editing and vice versa. We use a bin size of 20 for the division of the normalized $\Delta S$ value, then calculate the mean histogram in log scale among all tests. To simplify the analysis, we assume that the high-functional range of normalized $\Delta S$ for the feature editing result is higher than 0.6. Figure~\ref{fig:fig07} shows that the number of high-functional dimensions in $\Delta S$ is very low, and most of $\Delta S$ values are smaller than 0.6. The red dashed line indicates a count of 1, therefore histogram bars below it indicate dimensions for unused for many tests. The standard deviation is indicated as error bars on the graph, showing that the distribution is similar for all tests.

Based on the first experiment results, we then conduct the second experiment to explore the statistical convergence of $\Delta S$ value in the feature editing control task. For simplicity, the $\Delta S$ is a 4928 dimensional vector, we consider the top 50 absolute values of $\Delta S$ (See equation~\ref{eqn:eqn10}) drive the editing, and define the \textit{reuse rate} (See equation~\ref{eqn:eqn11}) to find out how frequently are they be used for each image editing test. 

\begin{equation}
    T = \{t \in \Delta S | \operatorname{rank_{\Delta S}}(t) \leq 50 \}
    \label{eqn:eqn10}
\end{equation}
where t is the dimension value of $\Delta S$, $\operatorname{rank_{\Delta S}}$ is the absolute value rank in $\Delta S$, rank 1 indicates the highest absolute value.

\begin{equation}
    R_{reuse}=\frac{card(x)}{N}, \ x \in X
    \label{eqn:eqn11}
\end{equation}
where $x$ is the element of set $X$, $N$ is the number of test, $card(x)$ is the cardinality of $x$ in the set $X$.
\vspace{0.5em}

As the style dimensions with the top 50 absolute values may vary in each experiment, we calculate the union set of corresponding dimension indices among all tests. Specifically, we conduct 10 feature editing tests and get 10 sets $T_1, T_2, T_3, ..., T_{10}$, then define the union set of all dimensions $T_u$ as: 
\begin{equation}
\begin{gathered}
    T_u = T_1 \cup T_2 \cup T_3 ... \cup T_{10}.
\end{gathered}
\label{eqn:eqn12}
\end{equation}

\begin{figure*}[htb]
\centering
    \includegraphics[width=1.0\textwidth]{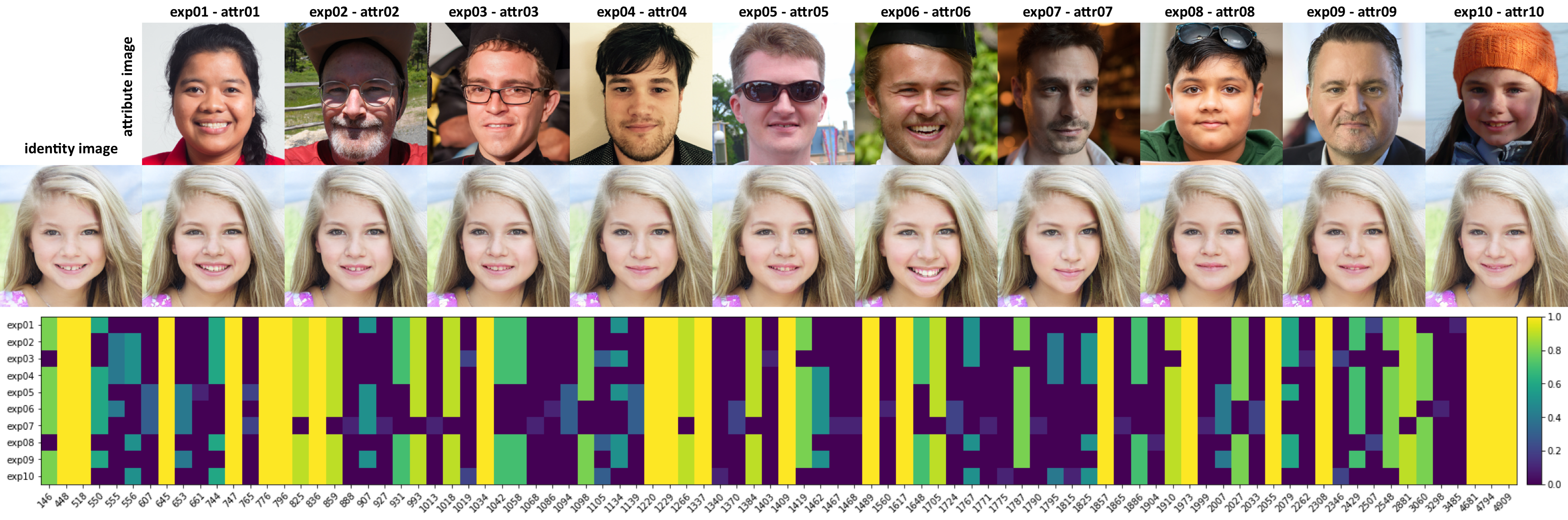}
    \caption{Statistics of channel reuse proportion for top 50 absolute control value of style space among all experiments, the yellow colour indicates the highest reuse rate $R_{reuse}$, the purple colour indicates the lowest reuse rate, the X-Axis is the channel index in the union set $A_u$ of top 50 control channels among all experiments.}
    \label{fig:fig08}
\end{figure*}

We draw the reuse rate $R_{reuse} $ distribution of the union set $T_u$ for each test in Figure~\ref{fig:fig08}. The top part of the figure shows the edited results where 10 different attribute images were drawn from the real faces dataset FFHQ256, and the identity image is from the synthetic dataset. The bottom panel illustrates the reuse rate $R_{reuse}$ of each dimension in $T_u$ for every test where a yellow colour indicates a higher reuse rate, purple colour indicates a lower reuse rate: $T_u$ contains 94 channels sorted in ascending order along the X-Axis. The 10 tests are arranged along the Y-Axis, where each row corresponds to each column in the top panel. This graph shows that editing an image for a range of pose and expression is driven by the same small number of style control signal dimensions (Sec.~\ref{term:stylecontrolsignal}), such as $\Delta S_{448}, \Delta S_{518}, \Delta S_{645}, \Delta S_{747}, \Delta S_{776}$ etc. 
Those few dimensions have most influence on the edited result among all tests, showing that the image editing of pose and expression for StyleGAN2 generator is mainly dependent on a few convolution channels whose activation can be controlled by the corresponding style control signal.

\vspace{-0.8em}

\subsubsection{Generative fields evaluation}
The theory of generative fields provides us with insights into the effect of editing features. Hypothetically, an editing signal can only control features that are smaller than the associated generative fields, for example, if the largest generative field is significantly smaller than the average face scale of the dataset, it cannot control the head movement due to insufficient influencing area. In our experiments, the average face size of FFHQ256 and our synthesized dataset is 141.68 pixels (we calculate it by using the \nth{1} and \nth{17} facial landmarks, left and right temples).

We design an experiment to test above hypothesis by disabling a set of style control vectors $\Delta S$ (Sec.~\ref{term:stylecontrolsignal}) according to the generative field size from its input feature map and evaluating the editing performance. Specifically, we calculate the mean metrics result from 100 tests where identity images are randomly sampled from our synthesised dataset and attribute images are randomly sampled from FFHQ256 dataset. The StyleGAN2 generator with $256\times256$ resolution has 13 convolution layers from $\texttt{conv0}$ to $\texttt{conv12}$ (Sec.~\ref{subsec:generativefields}), we define the control units in each experiment configuration that keep the style control signal injection for these convolution units but set others to $0$. For instance, the control units for \textbf{configuration~1} only keep $\texttt{conv0}$ to $\texttt{conv7}$ working for the feature editing, and set style control signal to $0$ for other convolution units, which covers generative fields size from 43 pixels to 506 pixels. Table~\ref{tab:tab03} compares the relationship between model performance and the generative fields (GFs) of the functional control units.

\begin{table}[h]
  \centering
  \resizebox{0.8\linewidth}{!}{
    \begin{tabular}{l l l l l l}
    \toprule
    Index & Control Units & GFs & Identity $\uparrow$ & Expression $\downarrow$ & Pose $\downarrow$ \\
    \midrule
    config.1 & conv0 - conv7   & (43, 506)  & 0.75 & 0.016 & 0.99 \\
    config.2 & conv0 - conv4   & (123, 506) & 0.74 & 0.019 & 1.04 \\
    config.3 & conv0 - conv2   & (251, 506) & 0.76 & 0.028 & 0.67 \\
    config.4 & conv3 - conv6   & (59, 187)  & 0.73 & 0.017 & 1.21 \\
    config.5 & conv6 - conv11  & (7, 59)    & 0.69 & 0.033 & 0.84 \\
    \bottomrule
    \end{tabular}
  }
  \caption{Quantitative evaluation of generative fields experiment}
  \label{tab:tab03}
\end{table}

\begin{figure}[h]
\centering
    \includegraphics[width=0.6\textwidth]{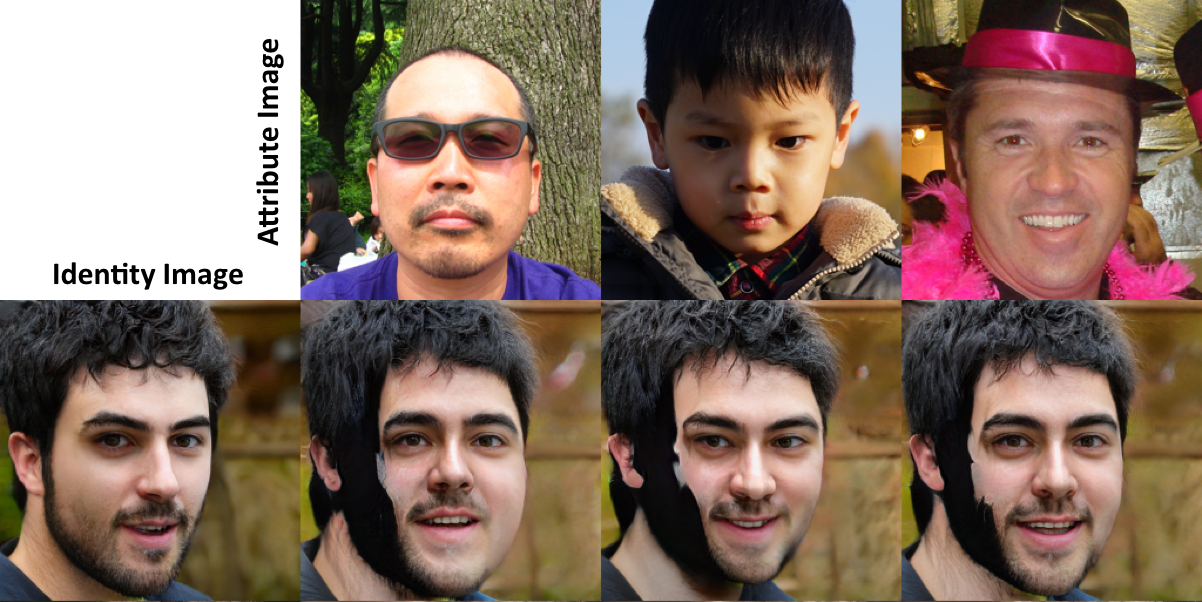}
    \caption{Small generative field works for the expression landmark editing, but the result gets the broken artifact due to limited influencing area, extremely decaying the image quality.}
    \label{fig:fig09}
\end{figure}

These results confirm our hypothesis: images produced from configuration~3, where the generative field is larger than the average face scale, lose control of finer-level features, like expression, but get the best result for controlling pose; results from configuration~5, where the generative field is significantly smaller than the average face scale, get the worst results of identity and expression for all experiments. We find that the pose result of configuration~5 is better than configuration~4 (shown in figure \ref{fig:fig09}) because the small generative fields cannot maintain the quality of the generated image, and the face is forced to match the head pose with many torn areas, which is reflected in the lowest identity and expression result.

\section{Conclusion}
We have presented a new method for fine-grained feature editing by using disentangled style space in StyleGAN2, demonstrating improved performance compared to existing methods. We also proposed a theoretical explanation for the feature control with different fineness levels of convolution units by introducing the inverted process of receptive field theory in generative models and analyzing the dependency issue of initial input of GANs, which are used to form the hypothesis and general editing approach. We discovered the style space has the sparse property to control a specific feature variation, and we use this to evaluate the generative field experiment.

\paragraph{Limitations} Our method does not consider full 3D supervision for reference images which means StyleGAN2 does not have enough information to guide the movement of the back half of the head; future work could use a 3D morphable model (3DMM) to replace the landmark reference to provide stronger 3D guidance.

\bibliographystyle{elsarticle-num-names} 
\bibliography{bibliography}






\end{document}


{\huge \textbf{Appendix:}}
\section{Generative fields size for all convolution layers}
Table \ref{tab:tab01} provides the generative field size of each convolution layer in style space $\mathcal{S}$, we calculate them by using the generative fields formula defined in section 3.2 of the paper. All convolution layers are indexed from \texttt{conv0} to \texttt{conv12} for the pre-trained StyleGAN2 generator with $256\times256$ resolution, we refer $\mathcal{S}$ layer index from \citet{wuStyleSpaceAnalysisDisentangled2021}, items that have same color in the table belong to the same generator block of StyleGAN2.
\vspace{0.5em}

\begin{table}[h]
    \centering
    \begin{tabular}{ccccc}
        input resolution & $\mathcal{S}$ layer index & conv layer index & generative fields & \# channels \\
        \rowcolor{lightgray}
        $4\times4$ & s0 & conv0 & 506 & 512 \\
        $8\times8$ & s2 & conv1 & 379 & 512 \\
        $8\times8$ & s3 & conv2 & 251 & 512 \\
        \rowcolor{lightgray}
        $16\times16$ & s5 & conv3 & 187 & 512 \\
        \rowcolor{lightgray}
        $16\times16$ & s6 & conv4 & 123 & 512 \\
        $32\times32$ & s8 & conv5 & 91 & 512 \\
        $32\times32$ & s9 & conv6 & 59 & 512 \\
        \rowcolor{lightgray}
        $64\times64$ & s11 & conv7 & 43 & 512 \\
        \rowcolor{lightgray}
        $64\times64$ & s12 & conv8 & 27 & 256 \\
        $128\times128$ & s14 & conv9 & 19 & 256\\
        $128\times128$ & s15 & conv10 & 11 & 128 \\
        \rowcolor{lightgray}
        $256\times256$ & s17 & conv11 & 7 & 128 \\
        \rowcolor{lightgray}
        $256\times256$ & s18 & conv12 & 3 & 64 \\
    \end{tabular}
    \caption{Generative fields size of each convolution layer in StyleGAN2}
    \label{tab:tab01}
\end{table}
\vspace{-1em}

\section{Comparison of image editing results}
We provide a comparison for the image editing results with different functional generative field size (GFs), the attribute images are randomly sampled from real world dataset FFHQ256, identity images are randomly generated from a pre-trained StyleGAN2 generator. As we mentioned in the paper (Sec.~4.2.2), the average face size of dataset is 141.68 pixels, figure \ref{fig:appendix_fig01} illustrates the comparison of feature editing results where smallest functional generative field sizes are below and above the average face size, it could obviously find that the model lose the expression feature control when the minimum generative field size is higher than the average face size.

\begin{figure}[htb]
\centering
    \includegraphics[width=0.85\textwidth]{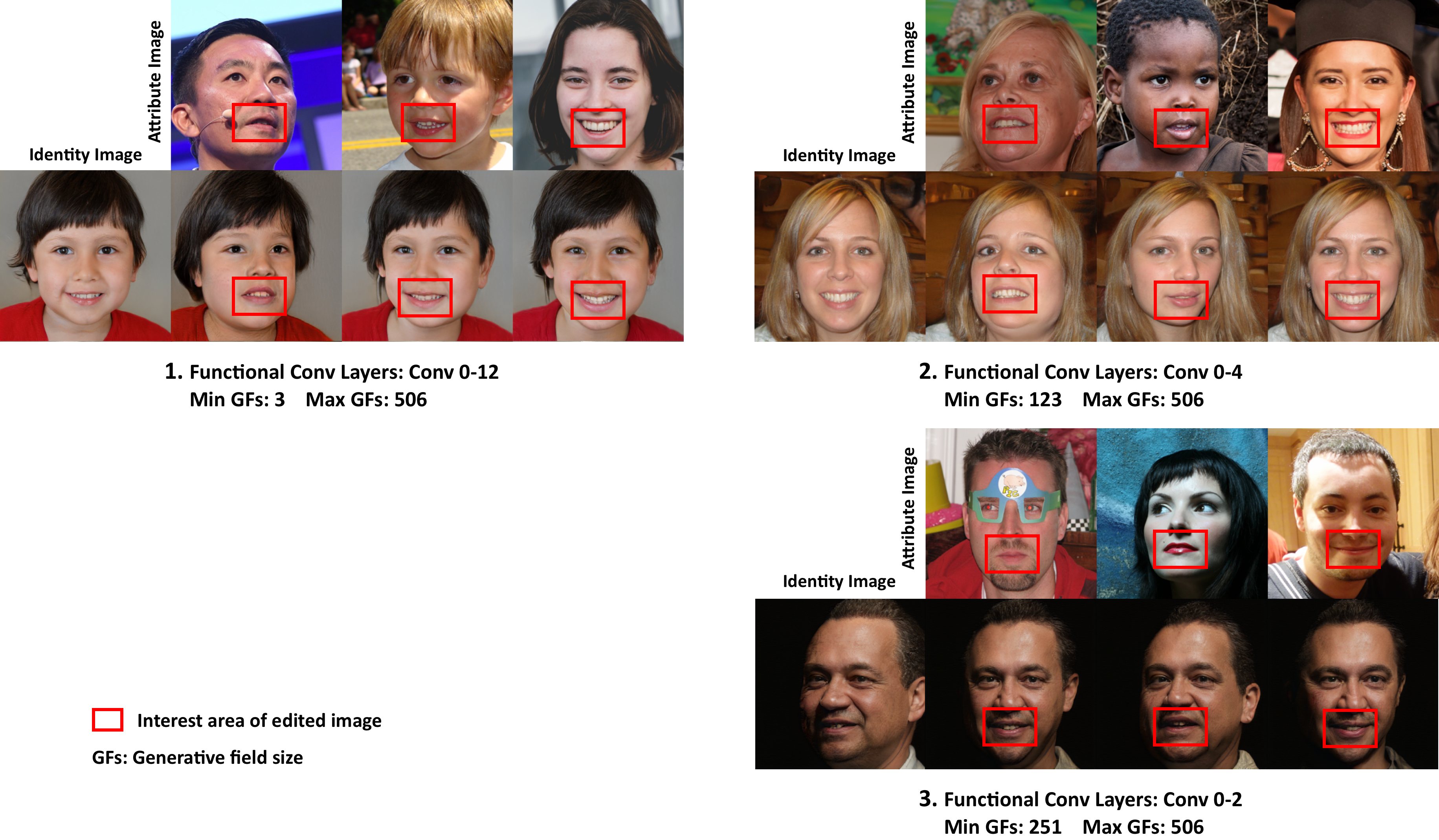}
    \caption{Image editing results comparison. \texttt{Image1} is the default setting which enables all GFs; the minimum GFs of \texttt{image2} is below the average face size; the minimum GFs of \texttt{image3} is above the average face size.}
    \label{fig:appendix_img01}
\end{figure}

We also compare image editing results which only use smaller generative fields. The largest generative field size of \texttt{image1} is slightly higher than the average face size which can reserve a little pose editing capacity with few torn artifact; for \texttt{image2}, the largest generative field size is lower than the average face size, it doesn't has enough influencing area to edit the whole head feature, then losing the head pose control.
\vspace{0.5em}

\begin{figure}[htb]
\centering
    \includegraphics[width=0.85\textwidth]{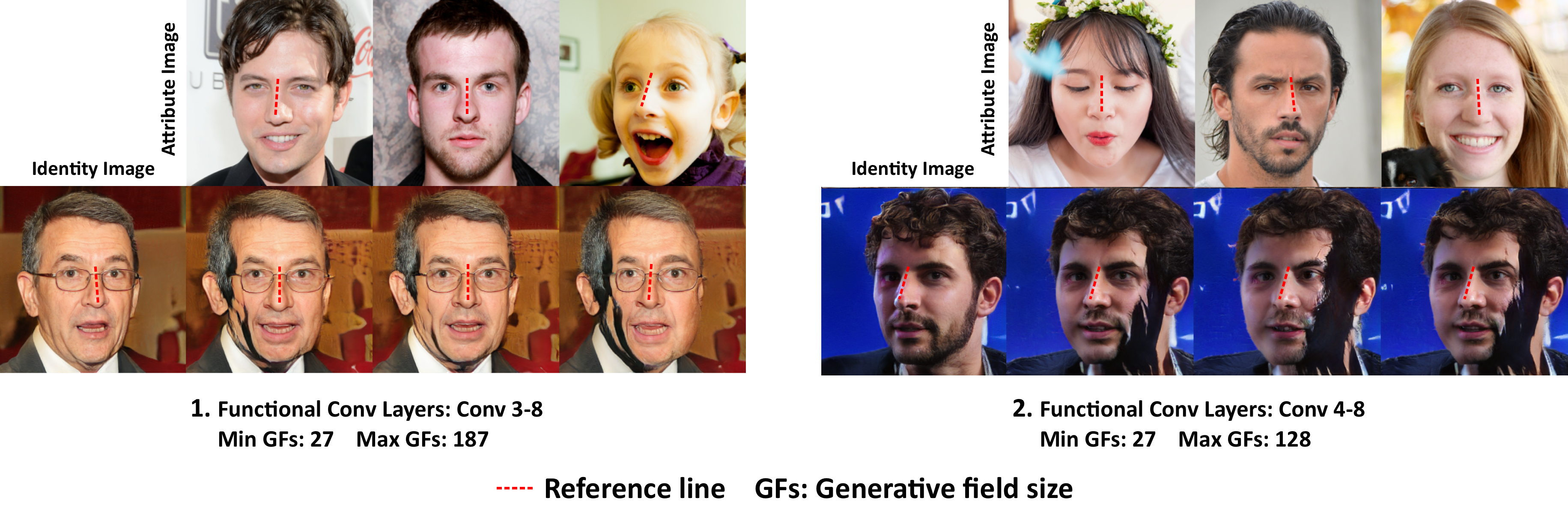}
    \caption{Image editing results comparison using limited generative field size.}
    \label{fig:appendix_img02}
\end{figure}

\begin{figure}[htb]
\centering
    \includegraphics[width=0.45\textwidth]{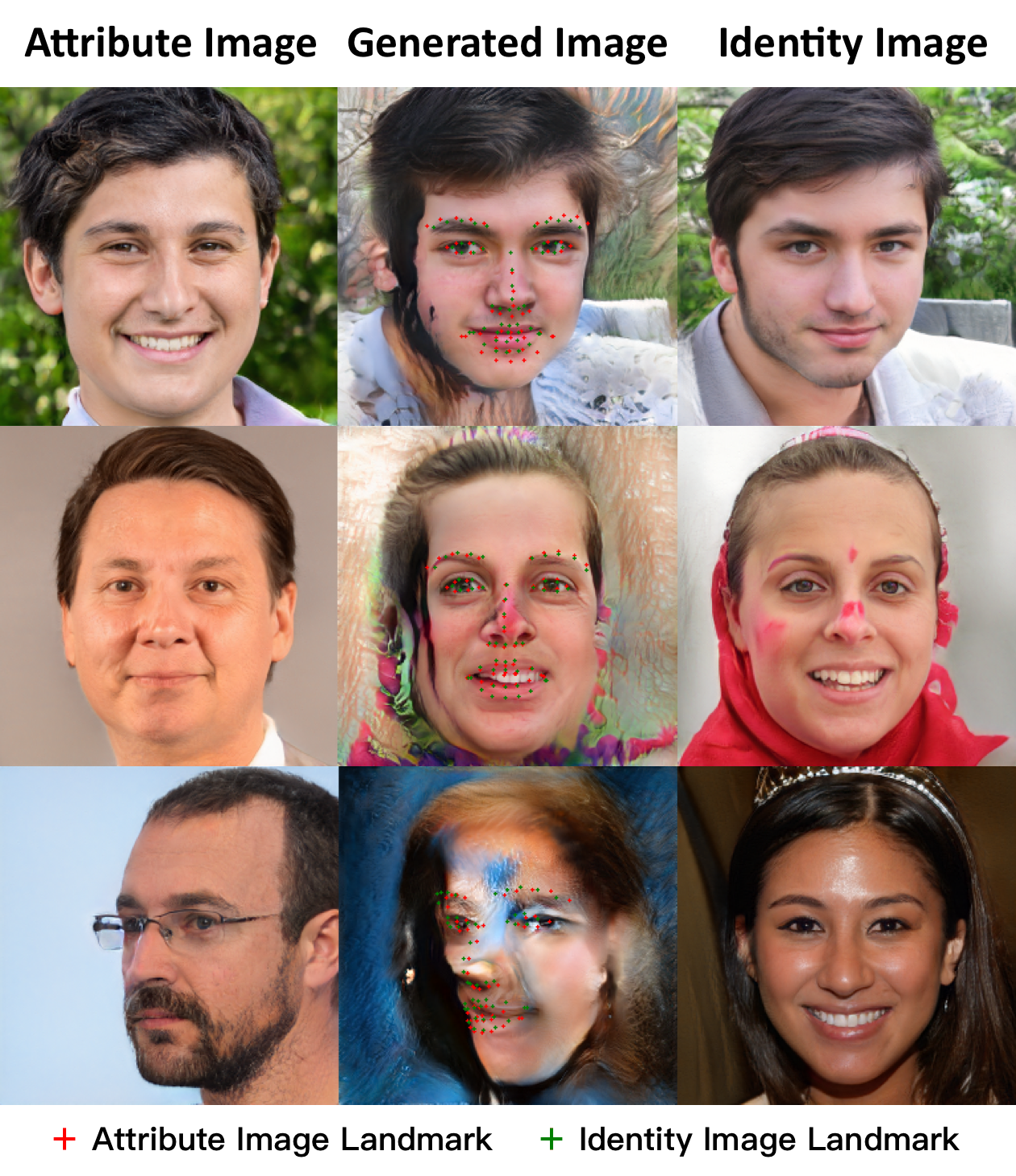}
    \caption{Image editing results with generative field size from 7 to 59.}
    \label{fig:appendix_img03}
\end{figure}

We find if only enabling very small generative field sizes the model would force to match the head pose and expression by downgrading the generated image quality (See figure \ref{fig:appendix_img03}), which could consist of many torn areas or pulled facial features. We explain it through it could cheat the pre-trained landmark detector to produce a matching result (See the overlap of detected landmark in figure \ref{fig:appendix_img03}).

\bibliographystyle{elsarticle-num-names} 
\bibliography{bibliography}